\documentclass[10pt,journal,compsoc]{IEEEtran}

\usepackage{amsmath}
\usepackage{amssymb,subfigure,cases}
\usepackage{color,hyperref}
\usepackage{algpseudocode}
\usepackage{booktabs} 

\usepackage{bbding}
\usepackage{amsmath}
\usepackage{subfigure}
\usepackage{caption}
\usepackage{algorithm}
\usepackage{algpseudocode}
\usepackage{setspace}

\usepackage{soul}
\usepackage{pifont}

\usepackage{times}
\usepackage{epsfig}
\usepackage{graphicx}
\usepackage{amssymb}
\usepackage{multirow}
\usepackage{subfigure}
\usepackage{pifont}
\usepackage{color,xcolor,colortbl}
\usepackage{array}
\usepackage{url}
\usepackage{CJK}
\usepackage{makecell}

\definecolor{mygray}{gray}{.9}

\newcolumntype{I}{!{\vrule width 1.2pt}}
\newlength\savedwidth
\newcommand\whline{\noalign{\global\savedwidth\arrayrulewidth
		\global\arrayrulewidth 1.25pt}%
	\hline
	\noalign{\global\arrayrulewidth\savedwidth}}

\ifCLASSOPTIONcompsoc
  \usepackage[nocompress]{cite}
\else
  \usepackage{cite}
\fi

\ifCLASSINFOpdf
\else
 \fi

\begin{document}
	
\setul{}{1.0pt}
%
\title{Counting Like Human: Anthropoid Crowd Counting on Modeling the Similarity of Objects}

\author{
        Qi~Wang, \IEEEmembership{Senior~Member,~IEEE,} Juncheng~Wang, Junyu~Gao,  Yuan~Yuan, \IEEEmembership{Senior~Member,~IEEE,}  Xuelong~Li, \IEEEmembership{Fellow,~IEEE} 
\IEEEcompsocitemizethanks{
\IEEEcompsocthanksitem 
Qi Wang, Junyu Gao, Yuan Yuan and Xuelong Li are with the School of Artificial Intelligence, Optics and Electronics (iOPEN), Northwestern Polytechnical University, Xi'an 710072, Shaanxi, China. E-mails: crabwq@gmail.com, gjy3035@gmail.com, y.yuan1.ieee@gmail.com, li@nwpu.edu.cn;\\
Juncheng Wang is with the School of Software and the School of Artificial Intelligence, Optics and Electronics (iOPEN), Northwestern Polytechnical University, Xi'an 710072, Shaanxi, China. E-mails: wangjunchengnwpu@gmail.com.

\textit{Xuelong Li is the corresponding author.}.
}}

\markboth{IEEE TRANSACTIONS ON PATTERN ANALYSIS AND MACHINE INTELLIGENCE,~Vol.~XXX, No.~XXX, XXX~XXX}%
{Shell \MakeLowercase{\textit{et al.}}: Bare Advanced Demo of IEEEtran.cls for IEEE Computer Society Journals}

\IEEEtitleabstractindextext{%
\begin{abstract}

The mainstream crowd counting methods regress density map and integrate it to obtain counting results. Since the density representation to one head accords to its adjacent distribution, it embeds the same category objects with variant values, while human beings counting models the invariant features namely similarity to objects. Inspired by this, we propose a rational and anthropoid crowd counting framework. To begin with, we leverage counting scalar as supervision signal, which provides global and implicit guidance to similar matters. Then, the large kernel CNN is utilized to imitate the paradigm of human beings which models invariant knowledge firstly and slides to compare similarity. Later, re-parameterization on pre-trained paralleled parameters is presented to cater to the inner-class variance on similarity comparison. Finally, the Random Scaling patches Yield (RSY) is proposed to facilitate similarity modeling on long distance dependencies. Extensive experiments on five challenging benchmarks in crowd counting show the proposed framework achieves state-of-the-art. 
\end{abstract}

}

\maketitle

\IEEEpeerreviewmaketitle

\section{Introduction}
\noindent Crowd counting aims to estimate the count of pedestrians within the image. The mainstream paradigm \cite{PCCNet, HYGNN, SASNet, DCECCV, cheng2022rethinking} for crowd counting is to regress a density map and integrate it to derive the counting result. The representation of density map tackles the sparse supervision signal of dots, in which the crowd counter is required to directly embed relationship between crowd semantic knowledge with density values. Hence, such a paradigm makes it feasible to pixel level regression. 

\begin{figure}
    \centering
    \includegraphics[width=0.5\textwidth]{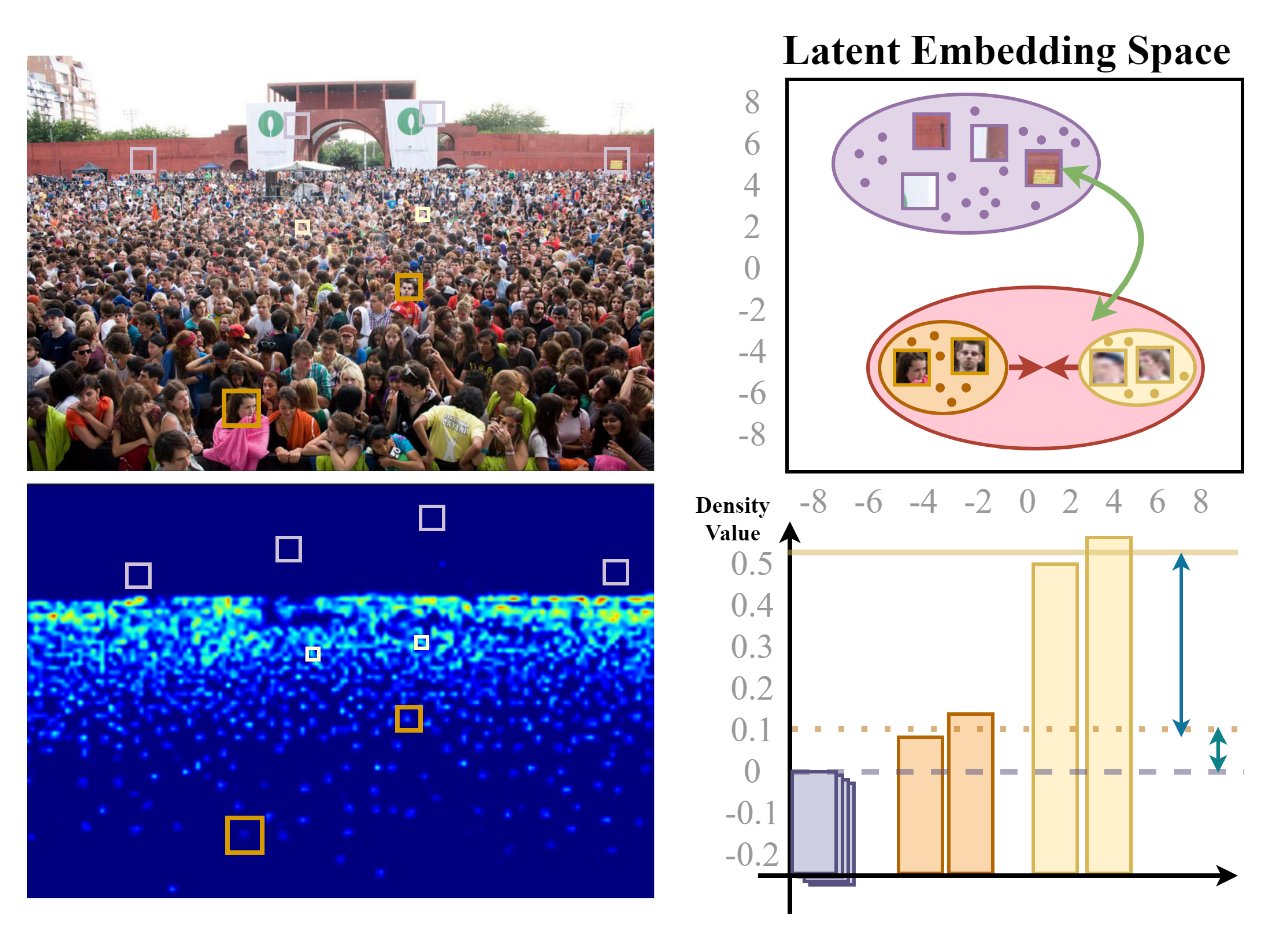}
    \caption{The ambiguity in density representation, in which the embedding relationship among heads and backgrounds are with inconsistence. Best view in color.}
    \label{Intro:density}
\end{figure}

Nevertheless, the density representation endows the same category objects namely pedestrians with variant density values according to the adjacent distribution. Concretely, as shown in Fig. \ref{Intro:density}, the semantic information of inner-classes, which are class \textit{pedestrians} and \textit{backgrounds}, should be embedded compactly. However, the density represents pedestrians with variant values, which even makes the decision boundary between pedestrian with background is closer than between pedestrians. Such a treatment is with ambiguity to human beings visual mechanism, in which the same category items are embedded based on the invariant features. Therefore, the counting for human beings is to compare the similarity among candidates, which endows us with strong generalization on counting, even the matters we never knew.

Inspired by this, we propose a rational and anthropoid framework on crowd counting. Our objective is to imitate human beings to absorb the invariant knowledge of the same category objects, then the counting paradigm is actually based on similarity comparison, not to rigidly model the mapping relationship between fixed semantic scenes with corresponding density values.

\begin{figure}[t]
    \centering
    \includegraphics[width=0.5\textwidth]{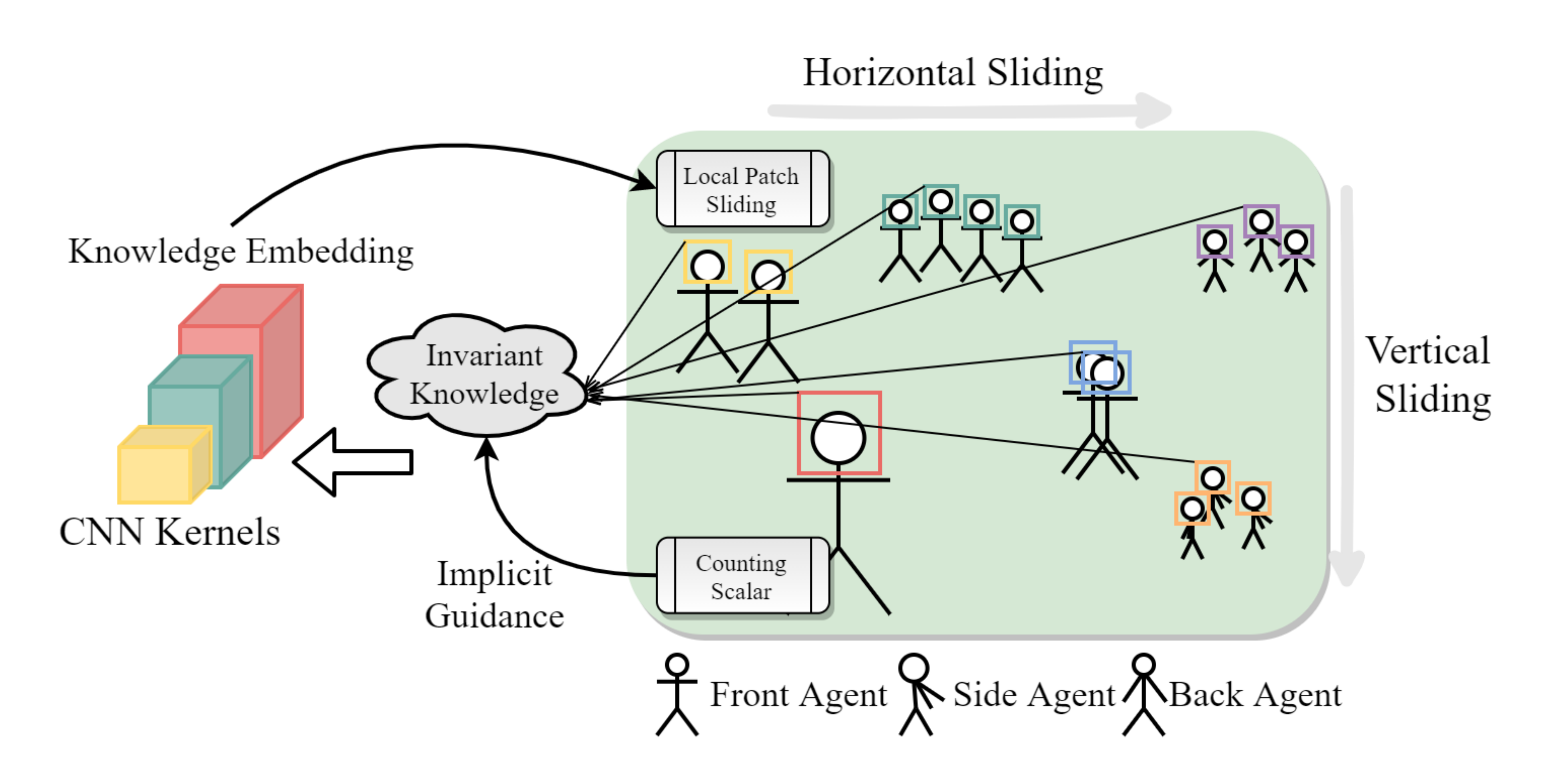}
    \caption{A toy example on our anthropoid crowd counting framework.}
    \label{toy}
\end{figure}

To begin with, the framework endeavors to learn the invariant knowledge in crowd scenes, and compare the similarity to derive the counting result. Thus, we disentangle the problem into invariant knowledge embedding and similarity comparison. A toy example can be found in Fig. \ref{toy}. Firstly, we leverage counting scalar as supervision signal to 
our framework. Comparing with density representation, the scalar signal is in global, which provides implicit guidance to the repeated and similar matters. Thus, the counter is inclined to concentrate on global invariant knowledge embedding, but not on local density mapping. Then, to better restore the invariant knowledge, the scalar supervision based Empirical Risk Minimization(ERM) should directly optimize the parameters. Hence, the CNN kernels are utilized to restore such invariant knowledge, and the parameters are updated via gradient. Secondly, with scalar based ERM optimizing kernels, the CNN is able to slide on the feature maps to compare the similarity among candidates with invariant knowledge, which is with analog to human beings. However, the conventional CNNs are limited on Receptive Field(RF). Despite that linearly stacking layers alleviates the issue, the Effective RF(ERF) is still limited \cite{ERFPaper} and the paradigm breaks the consistence among invariant knowledge with candidates. Thus, we present Large Kernel(LK) CNN to tackle the two issues, in which the LK is endowed huge ERF and promises the consistence among features.

 Albeit that the LK tackles the invariant knowledge modeling and promise the consistence among features, LK processes the knowledge in one view. Regarding to the crowd scenes, the variant features for inner-class impedes the similarity modeling, especially for scale variance. To address the variant issue for inner-class, we propose to parallel small kernels with large kernels in capturing the multi-scale knowledge. To this end, the invariant knowledge is the aggregation from multi-views and more general. However, the paralleled small kernels improves the orthogonality among features from branches, which is an inductive bias and incurs counter hard to converge. Inspired by \cite{RepVGG} and \cite{RepLKNet}, to facilitate model convergence without sacrificing variance modeling capacity for inner-class, we propose to structurally re-parameterize the pre-trained parameters on ImageNet \cite{ImageNet}, which is large enough to omit the inductive bias. 

Despite that the LK alleviates the limitation of CNN kernels on invariant knowledge modeling via enhancing the ERF, the CNN architecture cannot model the similarity on long distance dependencies. In crowd scenes, the objects with less inner-class variance tend to be distributed compactly in spatial. Therefore, the long distance dependencies facilitate region level similarity modeling. Comparing with CNNs, human beings address the long distance dependencies via skipped concentration windows, which can be described as the action of \textit{glance}. Nevertheless, we assert the skipped windows are hard to deploy, due to the limitation on existing CUDA operators. Thus, we propose Random Scaling patches Yield(RSY) to slide multi-scale windows from input perspective to model long distance similarity.

In a nutshell, our contributions can be four-fold:
\begin{itemize}
    \item Design an anthropoid framework on crowd counting to model the similarity of objects. The framework is a large kernel CNN supervised by counting scalar, in which the scalar supervision signal implicitly guide large kernel to model the similarity.
    \item Present to re-parameterize the pre-trained large kernel CNNs parameters to facilitate convergence without sacrificing the inner-class variance representation.
    \item Propose the Random Scaling patches Yield to model the similarity to long distance dependencies, which imitates the \textit{glance} to human beings.
    \item Demonstrate SOTA performance on five mainstream crowd counting datasets. To verify the similarity modeling, experiments on multi-class object counting achieves superior performance.
\end{itemize}

\section{Related Works}
\subsection{Crowd Counting}
The existing crowd counting methods can be divided into two main branches and other rare paradigms.
Firstly, density map regression\cite{WeizhePami, SASNet, AutoScale}  is to convolve the annotated dots with Gaussian kernel, then the pixel wise regression is implemented. The counting result accords to the integrating of the density map. However, as aforementioned, the density represents the same category objects with variant values, which is ambiguity with human visual mechanism. Moreover, the density map generation is cumbersome, for there requires too many manual settings.

Secondly, counting from localization can intuitively locate each instance and count. The object detection for crowd localization is proposed \cite{TinyFaces}. But the object detection paradigm needs complex post processing, like NMS, which neglects too many candidates. Thus, \cite{TopoCount, IIM, ScopedTeacher} introduce foreground segmentation into crowd localization, in which the heads are segmented as foreground. Nevertheless, localization based counter performs poorly under blurred regions for its low confidence to locate the objects.

Thirdly, there are some other paradigm crowd counters. \cite{Bayes, DMCount, MAN} utilize dots supervision. To overcome the sparsity of dots, the complex matching processing and strong assumption are implemented. Moreover, due to the limitation on annotation, some dots cannot be annotated in the middle of heads, which incurs noisy labels to such dot supervision. Then, the scalar supervised crowd counting \cite{TransCrowd, CrowdFormer, CrowdMLP} also exists. Since the scalar signal is in global and weak, the previous works concentrate on expanding model global attention with ViT or MLP, but neglects the essential and rational features in scalar.
\begin{figure}[t]
    \centering
    \includegraphics[width=0.5\textwidth]{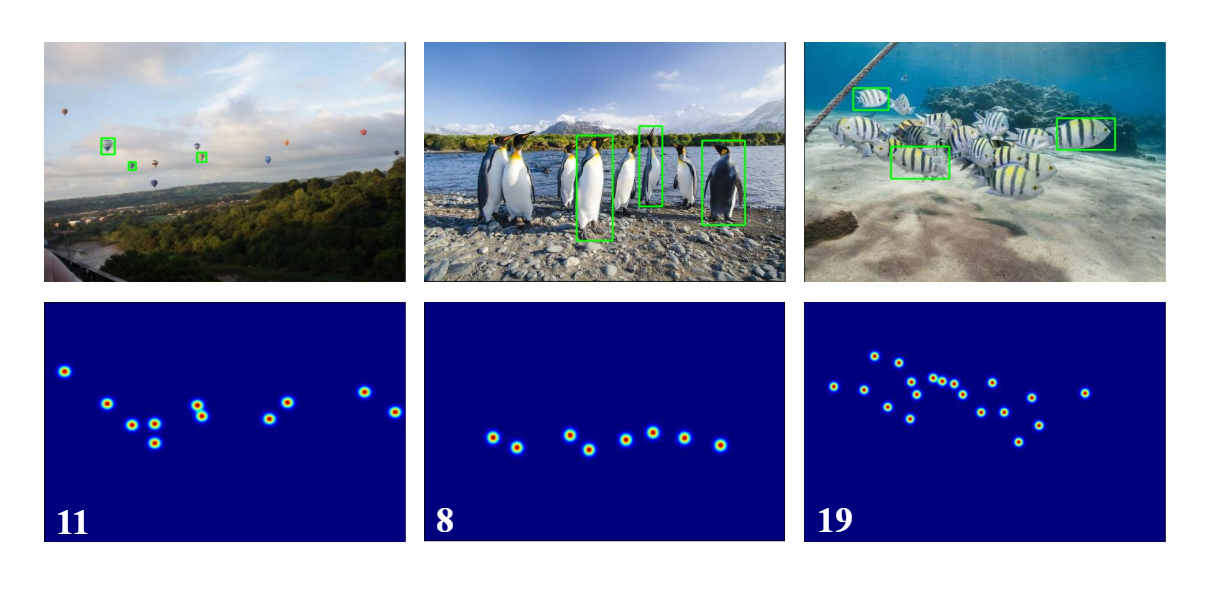}
    \caption{Typical samples in FSC-147, in which few objects are annotated via boxes and each one is annotated via a dot. To facilitate better illustration, the annotated dots are represented with Gaussian blur.}
    \label{FSC_box}
\end{figure}
\subsection{Similarity Counting}
The crowd counting only counts the pre-defined category namely pedestrians. To endow model capacity on counting class-agnostic objects, the similarity comparison is vital to learn the repeated matters. The mainstream paradigm for class-agnostic counters requires manual interaction to annotated few boxes to the objects of interest. The released benchmark FSC-147\cite{FamNet} is for such task,  in which a few boxes are annotated and other objects belonging to the same category are annotated via dots, as Fig. \ref{FSC_box} shown. Therefore, the previous works\cite{FamNet} adhere to the paradigm, which extracts invariant knowledge according to boxes and utilizes knowledge to compare similarity. To this end, the model can generalize to class-agnostic settings.
We assert that the paradigm is with analog to human visual mechanism, but it requires interactive guidance to invariant knowledge, while human beings not. Therefore, we further exploit an end-to-end invariant knowledge embedding.

\section{Our Approach}
This paper proposes a scalar crowd counting framework from the inspiration of human visual mechanism. To imitate the human beings counting which models invariant features firstly then slides to compare similarity, the large kernel CNN is leveraged. Then, the paralleled architecture and structural re-parameterization on pre-trained parameters are presented to enhance inner-class variance modeling and facilitate convergence. At last, the Random Scaling patches Yield (RSY) is proposed to aid the similarity representation on long distance dependencies.

\subsection{Anthropoid Counting Framework}
Human beings counting first models the invariant knowledge, and slides to compare similarity. Then, the counting result is from integrating the compared similar ones. Therefore, such a bottom-up paradigm endows human beings birthing with counting matters we never met before. To that effect, we propose an anthropoid counting framework imitating human beings counting. Hence, we disentangle the human beings counting into two parts, which are invariant knowledge embedding and similarity comparison. To this end, it is desirable that 1. supervision provides guidance to invariant knowledge and directly optimizes embedded knowledge; 2. knowledge is consistent within the feature to promise comparison meaningful.

\subsubsection{Invariant Knowledge Embedding}
As aforementioned, human beings counting models invariant knowledge, while the mainstream density represents each instance independently into discrete and different density values. Moreover, the density is actually an intermediate representation to our objective, due to the final counting result is from integrating the density map. Hence, the paradigm is with ambiguity and redundant to our objective. 

To this end, we exploit a human like supervision signal, in which the counting scalar is utilized as regression objective. Scalar signal is in global, which makes counter model the global dependencies but not concentrate on local information mapping. Thus, the global signal promises feature with global consistence, and the intra-class compactness is also constrained under such features. Hence, the scalar supervision provides implicit guidance to similar matters. Moreover, scalar regression is directly optimizing our final objective. With the counting scalar as supervision, the computational complexity of Empirical Risk Minimization(ERM) based optimization is deduced. 

Then, to better restore the guided knowledge by scalar, the CNNs with directly optimized kernels from ERM are leveraged. In optimization, the parameters in kernel are updated according to the gradient, which aims to deduce the counting scalar loss namely empirical risk. Hence, the scalar guided implicit invariant knowledge can directly optimize the kernels for better embedding.

\subsubsection{Generalized Consistence Similarity Comparison}
With scalar as regression objective to directly optimize the kernels, the shared weights in CNN are able to slide on the feature maps to compare the similarity to imitate human counting. Nevertheless, the conventional CNN with frequently adopted sizes of 1, 3, 5, 7 are limited in spatial. As a result, the scope for modeling the invariant knowledge is bound to be small. Later, the content for objects is with high variance, such as perspective and occlusion. Thus, small kernels with deficient parameters are inclined to under-fit on the variance. 

Regarding to the stacked CNNs, it is demonstrated that the RF is growing as  $\sqrt{n}$, while the ERF is shrinking as  $\frac{1}{\sqrt{n}}$, where the depth grows $n$ times \cite{ERFPaper}. Moreover, we assert the stacked small kernels break consistence of features, which makes candidate embedding and feature map comparison be processed on different stages.

To this end, we propose to utilize large kernel CNNs to tackle the issues. Firstly, the large kernel CNN expands the ERF, which makes the outliers in scale can be captured in one layer or adjacent layers. Therefore, it enhances the invariant knowledge consistence in sliding windows when comparing the similarity. Secondly, the large kernel brings adequate parameters on coping with the content variance. To avert the excessive parameters incurred by large kernel simultaneously, the depth-wise convolution is conducted.

\begin{figure}[t]
    \centering
    \includegraphics[width=0.5\textwidth]{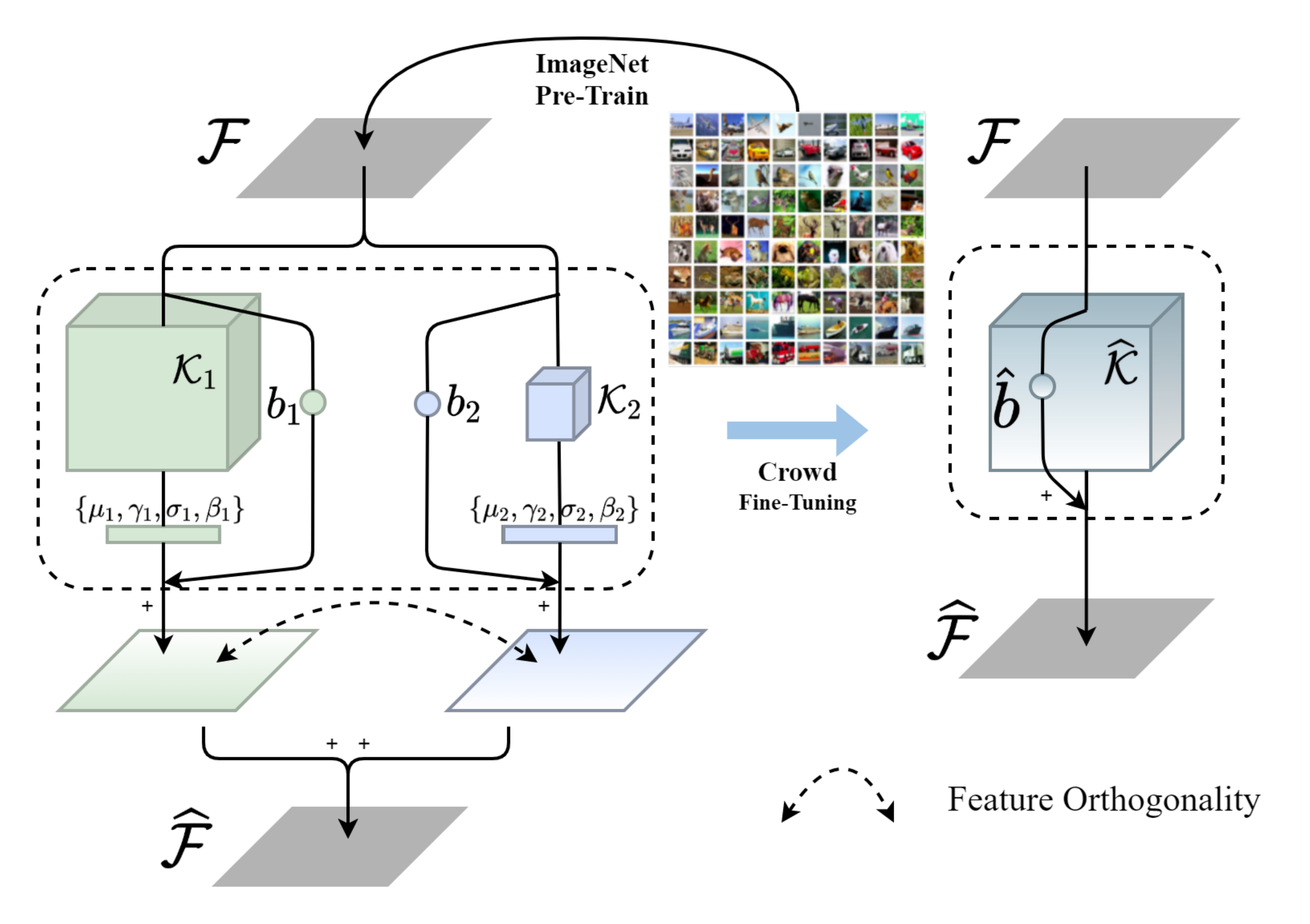}
    \caption{The pipeline of structural re-parameterization on pre-trained parameters.}
    \label{REP}
\end{figure}

\subsection{Scale Invariance Embedding}
With proposed large kernel CNN alleviating certain content variance embedding, the scale variance is still an essential issue for being ubiquitous in crowd scenes. Human eyes own strong scale invariance embedding capacity, while the fixed kernels and linearly stacked layers not. To enhance the capacity in embedding features with scale invariance, we propose to parallel layers using tiny kernels with the mainstream CNN flow. Significantly, we assert that the paralleled kernels should have untrivial size gap to enhance the orthogonality and compelmentarity between features extracted correspondingly. As shown in Fig. \ref{REP}, let us take an example with two branches. Given an input feature $\mathcal{F}$ and an output feature $\widehat{\mathcal{F}}$, the Eq. \ref{conv_f} can formally describe the process:
\begin{equation}
    \widehat{\mathcal{F}} = \sum_{br=1}^{2} BN(\mathcal{F}*\mathcal{K}_{br}, \left \{ \mu _{br} ,\sigma _{br},\gamma _{br}, \beta _{br}\right \} ),
    \label{conv_f}
\end{equation}
where the $BN$ denotes the BatchNormalization with parameters $\mu, \sigma, \gamma, \beta$. 
Polymerizing the above paralleled module, it endows model to capture multi-scale knowledge and be treated with more scale invariance. 

\subsubsection{Structural Re-Parameterization on Pre-Trained Parameters}
Albeit that the orthogonality and compelmentarity for paralleled branches endows multi-scale invariance modeling, it makes model hard to converge for the divergence of features, especially on smaller datasets with large variance, and the phenomenon can be deemed as inductive bias. Inspired by \cite{RepVGG} and \cite{RepLKNet}, to deduce the training difficulty on convergence without sacrificing multi-scale invariance modeling, we propose to pre-train the paralleled architecture on ImageNet\cite{ImageNet}, which is large enough to omit the inductive bias and implement structural re-parameterization before crowd counting.

To be concrete, let all parameters in Eq. \ref{conv_f} have been pre-trained on ImageNet. In Eq. \ref{conv_f}, the ideal training pipeline on crowd is depicted as Fig. \ref{REP}. Further Eq. \ref{infer} can also formally describe the process in crowd fine-tuning.
\begin{equation}
    \widehat{\mathcal{F}} = \mathcal{F} * \widehat{\mathcal{K}} +  \widehat{b}.
    \label{infer}
\end{equation}

Combining Eq. \ref{infer} and Eq. \ref{conv_f}, it is obvious that Eq. \ref{kernel} holds.

\begin{equation}
    \widehat{\mathcal{K}}_{br} = \frac{\gamma_{br}}{\sigma _{br}} \cdot \mathcal{K}_{br}, 
    \widehat{b}_{br}=-\frac{\mu _{br}\cdot\gamma _{br}}{\sigma _{br}} +\beta _{br},
    \label{kernel}
\end{equation}
where $br$ is to discriminate different branches, and the candidate set for $br$ is $\left \{ 1, 2 \right \} $ in Fig. \ref{REP}. Then, the two branches can be fused via Eq. \ref{final}:

\begin{equation}
    \widehat{\mathcal{F}} =  \mathcal{F} * \left [ \widehat{\mathcal{K}}_1\oplus pad(\widehat{\mathcal{K}}_2,  \widehat{\mathcal{K}}_1) \right ] + \sum_{br=1}^{2}\widehat{b} _{br},
    \label{final}
\end{equation}
where $pad(\mathbf{p}_1, \mathbf{p}_2 )$ is the padding process and denotes to pad $\mathbf{p}_1$ to the same shape with $\mathbf{p}_2$ via $0$. 

Right before fine-tuning on downstream dataset, the $\widehat{\mathcal{K} } ,\widehat{b}$ come from ImageNet pre-trained, formally by Eq. \ref{fuse}:
\begin{equation}
    \left \{\widehat{\mathcal{K} } ,\widehat{b}\sim p(\mathcal{K}_i, \mu_i ,\sigma_i,\gamma_i ,\beta_i|x_{ImageNet})|i=1, 2  \right \} .
    \label{fuse}
\end{equation}

Therefore, comparing with normal model initialization, the Eq. \ref{fuse} restores multi-scale information of ImageNet, in which the parameters are inclined to be scale invariance. With such initialized parameters, the fine-tuning on crowd datasets can inherit the capacity. What’ s more, the really updated parameters are tractable. In summary, the proposed re-parameterization on ImageNet endeavors on facilitating counting performance and convergence.

\subsection{Long Distance Dependencies Similarity}
With the large kernel CNN based counter presented, it remedies the limitation on ERF. Despite that the scope of invariant knowledge is expanded, the architecture of CNN still concentrates on local windows, but neglects long distance dependencies. Regarding to the crowd scenes, due to the perspective phenomenon, the objects with less inner-class variance tend to be distributed compactly in spatial. So that the long distance dependencies aid to model the region level similarity. With such a global guidance to crowd scenes, counting based on similarity modeling can rapidly match the similar regions. Thus, the bottom-up and coarse to fine similarity modeling paradigm is gradually implemented. 

As for human beings, the long distance dependencies modeling is endowed via the action \textit{glance}. Given a huge crowd scenes, human eyes are able to coarsely recognize the similarity among regions via jumping concentration, but not sliding windows. With the glance process on huge crowd scenes, human models global guidance which facilitates fine-grained invariant knowledge modeling and similarity comparison. To that effect, we make attempt to imitate glance to our framework.

As aforementioned, the concentration for CNN is limited within the window, which incurs the feature level of CNN hard to be global. Furthermore, the jumping concentration is inefficient to deploy via existing CUDA operators on GPU (the sliding strides for kernel are uncertain). To this end, we propose to treat the issue from data perspective. Comparing with sliding the CNN kernel window, the jumping sliding for images patches seems to be more effective. Thus, the proposed Random Scaling patches Yield (RSY) is such a strategy to the objective. 


To begin with,  we divide the image $ \mathcal{I}$ into a patch set $\mathcal{P}=\left \{ p_1^1,p_1^2,...,p_1^{N_w},...,p_{N_h}^{N_w} \right \}  $ with $N_h * N_w$ patches. To meet the scale uncertainty issue, the divided patches are with random scales. Then the Eq. \ref{shuffle} implements random 2D-shuffling for the set $\mathcal{P}$. 
\begin{equation}
    \mathcal{I} =shuffle_{2D}(\mathcal{P}).
    \label{shuffle}
\end{equation}
Later, the shuffled patches are utilized to recompose the image. To this end, the RSY shuffles the image patches to replace the jumping sliding window. Therefore, with RSY in training, the counter can model the long distance dependencies with the locality of CNN.

\subsection{Training Objective}
Summarizing the above proposed modules, our framework is to utilize the CNN based counter to directly regress the counting scalar to the input image. To this end, the training objective of our framework is formulated as Eq. \ref{obj}:

\begin{equation}
     \theta_{opt} \longleftarrow arg \min_\theta{ \mathcal{L}\left [ f_\theta(\mathcal{I} ), \mathcal{\tau }  \right ]},
    \label{obj}
\end{equation}

\noindent where $f_\theta$ is the counting model with parameters $\theta$, $\tau $ is the annotated counting scalar and $ \mathcal{L}$ can be any L1 based loss function. In this paper, the smooth L1 loss is adopted.

\section{Experiment}

\subsection{Dataset and Metrics}
To verify the effectness of our proposed framework, we conduct experiments of crowd counting on ShanghaiTech Part A $\&$ Part B (SHHA $\&$ SHHB)\cite{MCNN}, UCF-QNRF (QNRF)\cite{QNRF}, JHU-Crowd++ (JHU)\cite{JHU} and NWPU-Crowd (NWPU)\cite{NWPU}. To further show the superiority of our framework on actively learning the similar and repeated matters, we deploy our work on FSC-147\cite{FamNet} which is a multi-class objects counting benchmark. 

Following previous works on crowd counting or multi-class counting, the Mean Absolute Error (MAE) and root Mean Square Error (MSE) are adopted as the metrics, which are shown as Eq. \ref{MSE}:
\begin{equation}
    \notag MAE = \frac{1}{N}\sum_{n = 1}^{N}\left \| p_n-t_n \right \|^1_1,
    \label{MAE}
\end{equation}
\begin{equation}
    MSE = \sqrt{\frac{1}{N}\sum_{n= 1}^{N}\left \| p_n-t_n \right \|^2_2 },
    \label{MSE}
\end{equation}

\subsection{Implementation Details}
For backgrounds, a PyTorch framework of C3F\cite{c3f}, a GPU of NVIDIA Tesla V100 (32Gb version) and a random seed of 3,035 are utilized. For data pre-processing, a resolution alignment to the multiple of 384, a crop size of 384 and a probability of 0.5 for horizontal flipping are utilized. For network, a backbone model of RepLKNet34-L\cite{RepLKNet} and a Feature Pyramid Network (FPN) \cite{fpn} are utilized. Then, the regression head leveraged for scalar regression has been arrayed in Table. \ref{regression}. Concretely, the Multi-Layer Perceptron (MLP) is leveraged to project the compared features into counting scalar. To meet different input resolutions, the two dimensional Global Average Pooling(GAP) is utilized. For training, a batch size of 24, an optimization of Adam\cite{adam}, a loss function of smooth L1 and a learning rate of 1e-5 are utilized.

\begin{table}[h]
\centering
\renewcommand\arraystretch{1.2}
\begin{tabular}{c|c} 
\whline
Layer                    & Parameters           \\ 
\whline
Global Average Pooling   & (9, 9)               \\ 
\hline
Feature Flatten          & 1536 * 9 * 9         \\ 
\hline
Activation Layer 1       & ReLU                 \\ 
\hline
Multi-Layer Perceptron 1 & (1536 * 9 * 9, 128)  \\ 
\hline
Activation Layer 2       & ReLU                 \\ 
\hline
Monte Carlo Drop out     & 0.5                  \\ 
\hline
Multi-Layer Perceptron 2 & (128, 1)             \\
\whline
\end{tabular}
\caption{Detailed architecture for regression head.}
\label{regression}
\end{table}

\subsection{Analysis on Our Method}

\subsubsection{Ablation Study}
\begin{figure}[t]
    \centering
    \includegraphics[width=0.5\textwidth]{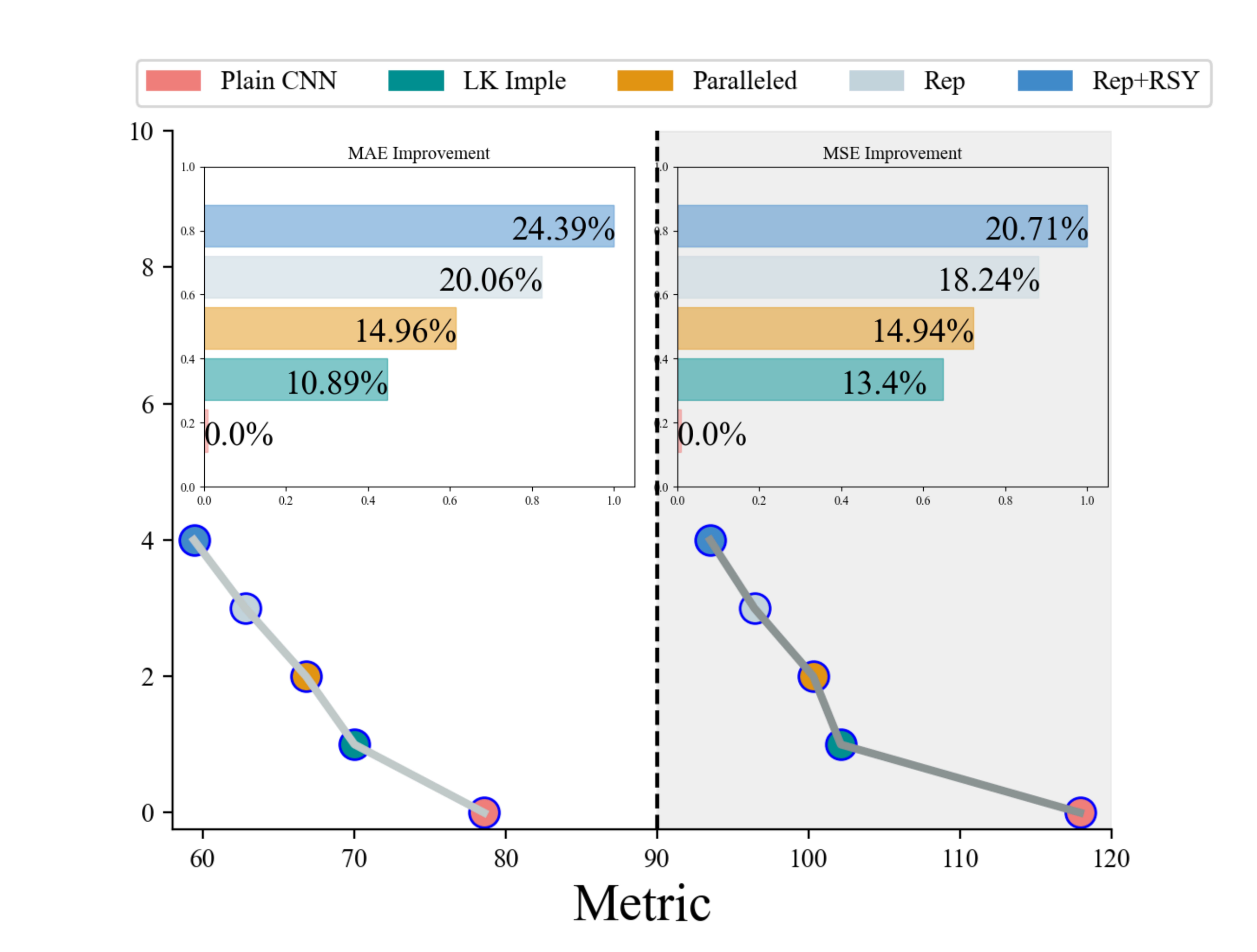}
    \caption{Ablation study on SHHA}
    \label{ablation}
\end{figure}

In this section, we decompose our proposed method into individual modules to see how each part contributes to the final result. The chosen dataset to implement ablation study is SHHA. The Fig. \ref{ablation} arrays the result.

\noindent\textbf{Plain CNN} which is our baseline model. We implement CNN model adopting usual kernels in VGG. \\
\textbf{LK Imple} In this item, we introduce large kernel into our model. As Fig. \ref{ablation} shown, the large kernel derives a promotion of 13.4\% on MSE and 10.89\% on MAE.
\textbf{Paralleled} Based on LK Imple, we parallel the small kernels along with the large kernel main branch. We find the certain promotion also obtained comapring with LK Imple.
\textbf{Rep} We structurally re-parameterize the small kernels trained on ImageNet. Hence, an untrivial promotion is achieved, in which 18.24\% on MSE and 20.06\% on MAE are obtained.
\textbf{RSY} In this subsection, we show the effectiveness on proposed RSY. The RSY further improves the MAE and MSE with 24.39\% and 20.71\% correspondingly.

\begin{table}[htbp]
\renewcommand\arraystretch{1.1}
\normalsize
\centering

\begin{tabular}{l|l|l|l|l} 
\whline
\multirow{2}{*}{Method} & \multicolumn{2}{l|}{LK Imple} & \multicolumn{2}{l}{VGG Imple}   \\ 
\cline{2-5}
                        & MAE           & ~MSE          & MAE           & MSE             \\ 
\whline
ScalarCrowd             & \textbf{59.4} & \textbf{93.5} & \textbf{69.9} & \textbf{104.2}  \\ 
\hline
DensityCrowd            & 67.1          & 113.8         & 71.5          & 117.6           \\
\whline
\end{tabular}
\caption{Rethinking different training paradigms on SHHA.}
\label{rethink}
\end{table}

\subsubsection{Scalar \textit{vs.} Density}
In this section, to show the superiority of scalar supervision, we make comparison between scalar with density under two settings. See Table. \ref{rethink}, we implement our whole framework namely ScalarCrowd in LK Imple 
\begin{table*}[t]
\renewcommand\arraystretch{1.1}
\centering

\begin{tabular}{c|c|c|c|c|c} 
\whline
\multirow{2}{*}{Method} & \multirow{2}{*}{Architecture} & JHU                        & SHHA                      & SHHB                         & QNRF\                           \\ 
\cline{3-6}
                        &                               & MAE/MSE                      & MAE/MSE                     & MAE/MSE                    & MAE/MSE                       \\ 
\whline
Sorting\cite{Sorting}                    & CNN                           & -/-                          & 104.6/145.2                 & 12.3/21.2                  & -/-                           \\ 
\hline
MATT\cite{MATT}                   & CNN                           & -/-                          & 80.1/129.4                  & 11.7/17.5                  & -/-                           \\ 
\hline
GLC Loss\cite{GLCLoss}               & CNN                           & 143.9/485.3                  & 81.2/127.1                  & 16.8/28.5                  & 161.9/273.7                   \\ 
\hline
GLC Loss\cite{GLCLoss}                 & ViT                           & 116.5 404.4~ ~               & 82.7/122.8                  & 14.9/25.5                  & 145.8/249.0                   \\ 
\hline
TransCrowd-T\cite{TransCrowd}             & ViT                           & ~ ~76.4/319.8                & 69.0/116.5                  & 10.6/19.7                  & 98.9/176.1                    \\ 
\hline
TransCrowd-G\cite{TransCrowd}               & ViT                           & 74.9/295.6                   & 66.1/105.1                  & 9.3/16.1                   & 97.2/168.5                    \\ 
\hline
CrowdFormer\cite{CrowdFormer}            & ViT                           & -/-                          & 62.1/94.8                   & 8.5/13.6                   & \underline{93.3}/\underline{160.9}    \\ 
\hline
Crowd-MLP\cite{CrowdMLP}             & MLP                           & \underline{67.6}/\underline{256.2}   & \textbf{57.8}/\textbf{84.4} & \underline{7.7}/\underline{12.1}   & 94.1/170.3                    \\ 
\hline
Ours          & CNN                           & \textbf{62.4}/\textbf{248.9} & \underline{59.4}/\underline{93.5}   & \textbf{7.4}/\textbf{11.4} & \textbf{90.9}/\textbf{158.8}  \\
\whline
\end{tabular}
\caption{Comparison with SOTA methods on JHU, SHHA, SHHB and QNRF. The compared methods are Sorting, MATT, GLC Loss, TransCrowd, CrowdFormer, Crowd-MLP. For fair comparison, we basically divide the methods according to their architectures. The bold text denotes the first place
and the underlined text denotes the second place.}
\label{NWPUSOTA}
\end{table*}
\begin{table*}[t]
\renewcommand\arraystretch{1.1}
\normalsize
\centering
\begin{tabular}{c|c|c|c|c|c} 
\whline
\multirow{3}{*}{Method} & \multirow{3}{*}{Architecture} & NWPU-V                                         & \multicolumn{3}{c}{NWPU-T}                                                                                                         \\ 
\cline{3-6}
                        &                               & Overall                                        & Overall                                        & \multicolumn{2}{c}{Scene Level (MAE only)}                                        \\ 
\cline{3-6}
                        &                               & MAE / MSE / NAE                                & MAE / MSE / NAE                                & Avg.           & S1-S4                                                            \\ 
\whline
GLC Loss               & CNN                           & 138.6 / 512.2~/ -                              & 159.1 / 508.4~/ -                              & -              & - / - / - / -~                                                   \\ 
\hline
GLC Loss              & ViT                           & 135.4 / 512.0~/ -                              & 137.4 / 425.6~/ -                              & -              & - / - / - / -~                                                   \\ 
\hline
TransCrowd-T            & ViT                           & 88.2 / 446.9~/ -                               & 119.6 / 463.9 / -                              & 736.0          & 12.7 / 47.2 / 311.2 / 3216.1~ ~                                  \\ 
\hline
TransCrowd-G             & ViT                           & 88.4 / 400.5~/ -                               & 117.7 / 451.0 / 0.244                          & 737.8          & 12.8 / 46.0 / 309.0 / 3252.2                                     \\ 
\hline
Ours          & CNN                           & \textbf{56.9~}/~\textbf{179.6 }/ \textbf{0.12} & \textbf{86.7 }/ \textbf{435.0 }/ \textbf{0.17} & \textbf{641.5} & \textbf{8.7} / \textbf{34.6 }/ \textbf{199.3 }/ \textbf{2313.2}  \\
\whline
\end{tabular}
\caption{Comparison with SotA methods on NWPU-Crowd, in which the testing labels are agnostic to the public. The compared methods are GLC Loss\cite{GLCLoss}  and TransCrowd\cite{TransCrowd} . The S1-S4 denotes the number of pedestrians within the scene is in the interval of the sequence with landmarks 0, 100, 500, 5000 and postive infinite.}
\label{TabSOTA}
\end{table*}
\noindent column. Moreover, we concatenate a decoder module in \cite{c3f} to regress density map namely DensityCrowd in LK Imple column. Then, we replace our framework into a plain VGG-16\cite{VGG} namely VGG Imple columns.

As Table. \ref{rethink} shown, the scalar supervised ScalarCrowd achieves the first place on MAE and MSE. Comparing with DensityCrowd, there is a MAE comparison of 59.4 vs. 67.1 and a MSE comparison of 93.5 vs. 113.8. The ScalarCrowd surpasses DensityCrowd with margins of 13$\%$ on MAE and 22$\%$ on MSE. 

Moreover, we conduct detail comparisons between density with scalar supervised training. As shown in Table. \ref{density}, the detailed information on training time, convergence epoch, training memory on GPU and inference Flops.

\begin{table}[h]
\centering
\renewcommand\arraystretch{1.2}
\begin{tabular}{c|c|c} 
\whline
Items             & Scalar  & Density  \\ 
\whline
Training Time     & 3.563 s & 4.926 s  \\ 
\hline
Convergence Epoch & 66      & 208      \\ 
\hline
Training Memory   & 16.6 Gb & 20.6 Gb  \\ 
\hline
Inference Flops   & 2.19 G  & 19.9 G   \\
\whline
\end{tabular}
\caption{Comparison between density with scalar on other items.}
\label{density}
\end{table}
Concretely, the training time is measured under an iteration with a batch size of 24 and an input resolution of 384 * 384. The convergence epoch denotes epoch model achieves best results on validation of SHHA \cite{MCNN}. Training memory is with the same input settings to training time measurement. Inference Flops is only for task head. 

As the Table. \ref{density} shown, the scalar framework is easier and faster to train comparing with density. 

\subsection{Comparison with State-of-the-Arts}

\subsubsection{Comparison on Fully Supervised Counting}

In this section, we compare our proposed framework with other methods adopting the same training paradigm which is scalar supervision. Considering the above narrations, the Vision Transformer (ViT) and Multi-Layer Perceptron (MLP) namely non-local based counter could perform better than CNN namely local based with scalar representation to the crowd scenes. Therefore, to prominent our superiority, we divide the other methods into two clusters which are CNN based and non-CNN based. In CNN cluster, the chosen methods are Sorting, MATT and GLC Loss CNN version in which the conventional stacked convolution layers are deployed. In non-CNN cluster, the chosen methods are GLC Loss ViT version, TransCrowd with two versions (TransCrowd-T and TransCrowd-G), CrowdFormer and a MLP based CrowdMLP. We compare the counting results (see Table. \ref{TabSOTA} and \ref{NWPUSOTA}) on JHU, NWPU validation set (NWPU-V) and NWPU testing set (NWPU-T), SHHA, SHHB and QNRF with norms of MAE and MSE. Due to the lack of records on other datasets, the NAEs are only arrayed on NWPU-V and NWPU-T.

As shown by Table. \ref{TabSOTA}, our proposed framework achieves the first place on the norms of MAE and MSE, except on SHHA in which a rank of $2^{nd}$ is still obtained. Significantly, comparing with other best CNN counters, our LKCrowd surpasses in an untrivial margin (a MAE of 59.4 vs. 80.1 on SHHA and a MAE of 90.9 vs. 161.9 on QNRF). Moreover, in NWPU which is the most challenging benchmark, we outperform TransCrowd-T with a MAE comparison of 56.9 vs. 88.2 on validation set and outperform TransCrowd-G with a MAE comparison of 84.7 vs. 117.7 on testing set.

\subsubsection{Comparison on Multi-Class Object Counting}
As aforementioned, one of our motivations to the proposed framework is to imitate human visual mechanism on modeling the similar matters. To this end, the paradigm endows us capacity on counting things we never met before. Summarizing the characteristic, we conduct our framework on a multi-class objects counting benchmark. In multi-class objects counting scenarios, the counter is required to count the similar and repeated matters. The mainstream paradigm in training needs few annotation boxes and all points to each objects as Fig. \ref{FSC_box} shown. To the best of our knowledge, we are the first to attempt multi-class object counting without the facilitating of few boxes, and the spatial annotations.

\begin{figure}
    \centering
    \includegraphics[width=0.5\textwidth]{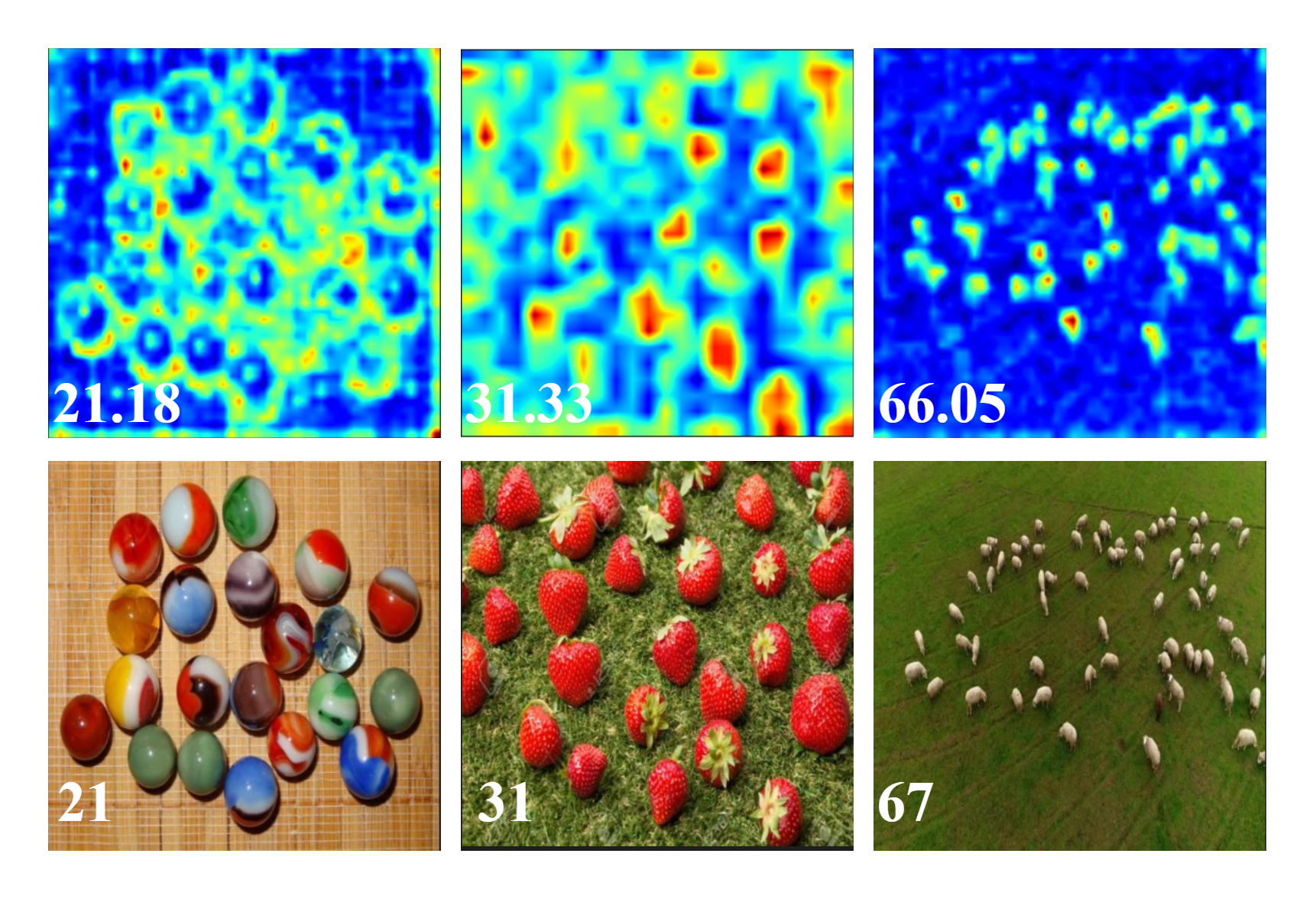}
    \caption{Typical images in FSC-147 are arrayed in bottom row. Significantly, the objects among class are with strong variance, even the class gap also exists between training set with validation set and test set. The visualized feature maps via channel-wise average pooling are arrayed in up row to show the effectiveness on actively model the similar ones.}
    \label{fsc}
\end{figure}

To be specific, the main benchmark to multi-class object counting is FSC-147, in which the 147 classes are included in the dataset, and some typical images have been depicted as Fig. \ref{fsc}. Therefore, with the scalar representation as supervision and objective, the performance on multi-class profoundly and directly demonstrates the reasonable and effective proposal to our framework. 
\begin{table}[htbp]
\centering
\renewcommand\arraystretch{1.1}
\normalsize
\begin{tabular}{c|l|c|c} 
\whline
\multirow{2}{*}{Methods} & \multirow{2}{*}{Interaction} & Validation                      & Test                     \\ 
\cline{3-4}
                         &                              & MAE / MSE                       & MAE / MSE                \\ 
\whline
FR~                      & \multicolumn{1}{c|}{\Checkmark }                           & 45.45 / 112.53                  & 41.64 / 141.04           \\ 
\hline
FSOD~                    & \multicolumn{1}{c|}{\Checkmark }                             & 36.36 / 115.00                  & 32.53 / 140.65           \\ 
\hline
MAML                     & \multicolumn{1}{c|}{\Checkmark  }                            & 25.54 / 79.44                   & 24.90 / 112.68           \\ 
\hline
GMN                   &\multicolumn{1}{c|}{\Checkmark }                            & 29.66 / 89.81                  & 26.52 / 124.57   \\ 
\hline
FamNet                   &\multicolumn{1}{c|}{\Checkmark }                            & 23.75 / \textbf{69.07}                   & 22.08 / \textbf{99.54}   \\ 
\hline
Ours                     &\multicolumn{1}{c|}{\XSolidBrush   }                             & \textbf{22.79} / 85.92                   & \textbf{18.72 }/ 122.41  \\
\whline
\end{tabular}
\caption{Comparison with SOTA methods on FSC-147. The compared methods are FR\cite{FR}, FSOD\cite{FSOD}, GMN\cite{GMN}, MAML\cite{MAML}, FameNet\cite{FamNet}.}
\label{FSC}
\end{table}

\begin{table}
\centering
\renewcommand\arraystretch{1.1}
\normalsize
\begin{tabular}{c|c|c|c}

\whline
\multirow{2}{*}{Method} & \multirow{2}{*}{Backbone} & B$\rightarrow $A                & A$\rightarrow $B               \\ 
\cline{3-4}
                        &                           & MAE /~MSE                       & MAE / MSE                      \\ 
\whline
D-ConvNet               & CNN                       & 140.4 /~226.1                   & 49.1 /~99.2                    \\ 
\hline
TransCrowd              & ViT                       & 141.3 /~258.9                   & 18.9 /~31.1                    \\ 
\hline
CrowdFormer             & ViT                       & 121.6 /~ 208.8   & 16.0 /~26.0  \\ 
\hline
Ours                    & CNN                       & \textbf{117.6 /~\textbf{207.3}} & \textbf{13.5 /~\textbf{21.8}} \\
\whline
\end{tabular}
\caption{Comparison with SOTA methods on Domain Gap.}
\label{Domain}
\end{table}
The detailed experimental results are arrayed in Table. \ref{FSC}. Significantly, our framework adopting scalar representation achieves comparative counting performance.We are only motivated by demonstrating the framework is reasonable on actively learning the similar matters. Nevertheless, we notice that our MSE is poor, which is incurred by outliers according to the sensitivity of MSE. We will discuss the phenomenon in the next.

To be specific, to the best of our knowledge, we are the first to implement multi-class counting without box interaction. Such a without interaction process can effectively demonstrate the capacity on actively learn the repeated matters, especially there is class domain gap between training set with testing set. However, we are wondering how it will work when there are more than one classes matters within one image. 

As shown in Fig. \ref{manual}, we manually concatenate different classes together to see how the model will output. Moreover, we visualize the feature map in the end of first stage. Then, we notice that the model pay attention to two classes and the predicted results is approximately equal with the sum of two classes. We deem that the phenomenon is incurred by implicit knowledge aggregation in MLP layers. To be concrete, the class A and class B are captured in different layers along the depth. In these corresponding layers, the candidates are decided to be similar. Since there is no guidance to the class of interest, MLP aggregates the two class similarity into one. Thus, the final result is the sum of two classes. 
\begin{figure}[htbp]
    \centering
    \includegraphics[width=0.45\textwidth]{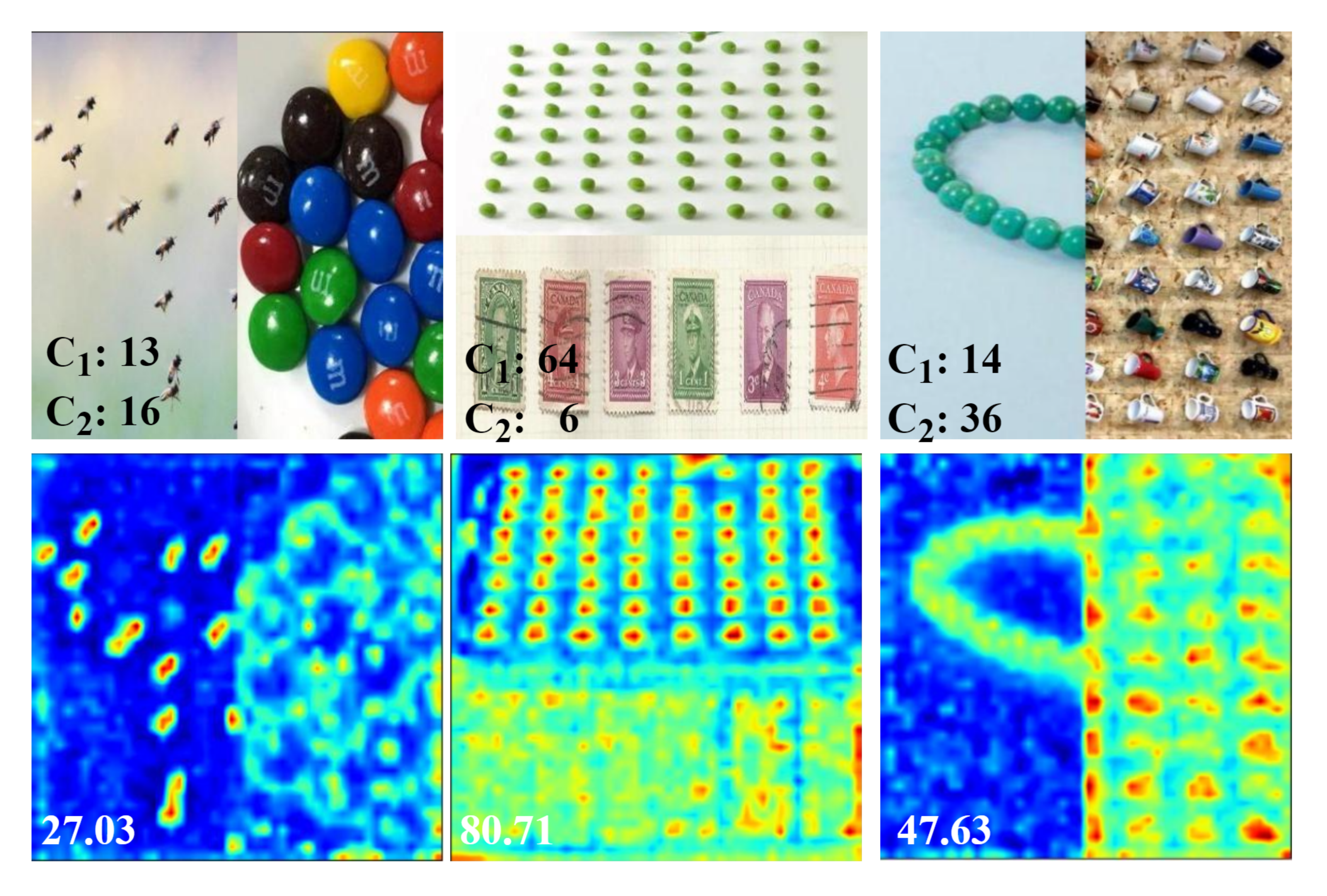}
    \caption{Manually concatenated samples on multi-classes samples. We visualize the feature map at the end of first stage via channel wise global average pooling.}
    \label{manual}
\end{figure}

Based on the analysis, we make attempt to explain the result in ablation study. As shown, our framework achieves a superior MAE but a poor MSE. This is inconsistency to other methods with box interaction. We dissert the crux to it. According the metrics computation, MSE is more fragile to outliers. Hence, the poor MSE denotes that our framework performs extremely poor on some samples. To further exploit the reason, we visualize some samples with extremely poor performance, as shown in Fig. \ref{manual}. We notice that the arrayed images exist class blurred phenomenon. The annotations indicate class A, while our framework concentrates on class B. Therefore, the performance outliers are generated and influencing the MSE of our framework.
\begin{figure}[t]
    \centering
    \includegraphics[width=0.45\textwidth]{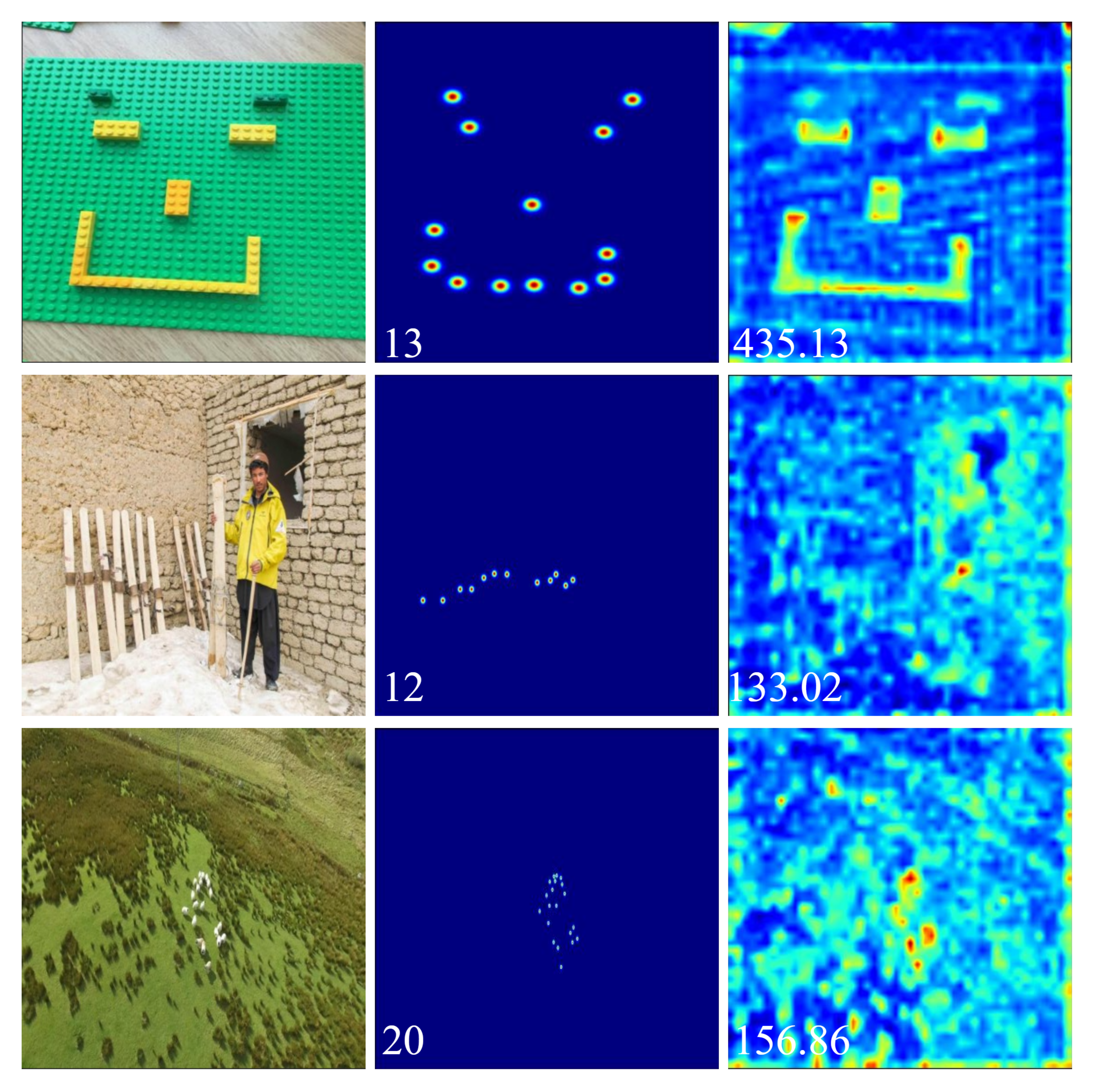}
    \caption{Typical failed samples on FSC-147 of our framework.}
    \label{manual}
\end{figure}

\subsubsection{Comparison on Domain Gap}
In this section, we discuss the performance on settings with domain gap, in which the training domain and testing domain are in distinction. The domain gap is a usual dilemma in real scenarios. As aforementioned, we deem our proposed framework is to imitate human beings, which should be generalized well. Thus, we deploy domain gap setting to verify the generalization of proposed method.

In our implementation, we pick SHHA and SHHB as two domains, in which the average density between two domains are ranked as $SHHA > SHHB$ . Thus, the density features among them exist domain gap. To this end, we implement the experiments as following. The $x\rightarrow y$ denotes the counter is trained on training set of $x$ and testing set of $y$. 

As the Table. \ref{Domain} shown, our framework surpasses the other ViT based scalar counter. Theoretically, the inference of ViT accords to the input self-attention, which means it is an adaptive scheme, while the CNN has the fixed parameters to cater to input with serious domain gap. Thus, when the input samples have domain gap to training samples, there is more challenge for CNN, and the architecture of ViT generalizes better under the domain gap. Nevertheless, our framework is based on similarity modeling which endows us more rationality, it is demonstrated the proposed framework has superiority. We still outperform ViT based scalar counter on mostly domain gap setting.

\section{Conclusion}
In this paper, we propose a rational and anthropoid framework on counting crowds. From the inspiration on human beings counting which first models invariant knowledge and compares similarity for counting, we leverage counting scalar to supervise the framework, which provides implicit guidance on similar matters. Moreover, we propose large kernel CNN along with structural re-parameterization on ImageNet pre-trained parameters based model to facilitate invariant knowledge modeling. Besides, the Random Scaling patches Yield is proposed to aid long distance dependencies similarity modeling. Extensive experiments on five crowd counting benchmarks demonstrate the superiority of our framework. What' s more, the experiment on multi-class object counting benchmark show the rationality of our framework.

In a word, this paper is motivated by human visual mechanism and proposes a baseline method for similarity modeling crowd counting. Therefore, we hope this paper could attract more research attention on counting with similarity modeling. 

{\small
\bibliographystyle{IEEEtran}
\bibliography{IEEEabrv,ref}

\begin{thebibliography}{10}
\providecommand{\url}[1]{#1}
\csname url@samestyle\endcsname
\providecommand{\newblock}{\relax}
\providecommand{\bibinfo}[2]{#2}
\providecommand{\BIBentrySTDinterwordspacing}{\spaceskip=0pt\relax}
\providecommand{\BIBentryALTinterwordstretchfactor}{4}
\providecommand{\BIBentryALTinterwordspacing}{\spaceskip=\fontdimen2\font plus
\BIBentryALTinterwordstretchfactor\fontdimen3\font minus
  \fontdimen4\font\relax}
\providecommand{\BIBforeignlanguage}[2]{{%
\expandafter\ifx\csname l@#1\endcsname\relax
\typeout{** WARNING: IEEEtran.bst: No hyphenation pattern has been}%
\typeout{** loaded for the language `#1'. Using the pattern for}%
\typeout{** the default language instead.}%
\else
\language=\csname l@#1\endcsname
\fi
#2}}
\providecommand{\BIBdecl}{\relax}
\BIBdecl

\bibitem{PCCNet}
J.~Gao, Q.~Wang, and X.~Li, ``Pcc net: Perspective crowd counting via spatial
  convolutional network,'' \emph{IEEE T-CSVT}, pp. 3486--3498, 2019.

\bibitem{HYGNN}
A.~Luo, F.~Yang, X.~Li, D.~Nie, Z.~Jiao, S.~Zhou, and H.~Cheng, ``Hybrid graph
  neural networks for crowd counting,'' in \emph{AAAI}, 2020, pp.
  11\,693--11\,700.

\bibitem{SASNet}
Q.~Song, C.~Wang, Y.~Wang, Y.~Tai, C.~Wang, J.~Li, J.~Wu, and J.~Ma, ``To
  choose or to fuse? scale selection for crowd counting,'' in \emph{AAAI},
  2021, pp. 2576--2583.

\bibitem{DCECCV}
H.~Xiong and A.~Yao, ``Discrete-constrained regression for local counting
  models,'' \emph{ECCV}, 2022.

\bibitem{cheng2022rethinking}
Z.-Q. Cheng, Q.~Dai, H.~Li, J.~Song, X.~Wu, and A.~G. Hauptmann, ``Rethinking
  spatial invariance of convolutional networks for object counting,'' in
  \emph{CVPR}, 2022, pp. 19\,638--19\,648.

\bibitem{ERFPaper}
W.~Luo, Y.~Li, R.~Urtasun, and R.~Zemel, ``Understanding the effective
  receptive field in deep convolutional neural networks,'' \emph{NeurIPS},
  vol.~29, 2016.

\bibitem{RepVGG}
X.~Ding, X.~Zhang, N.~Ma, J.~Han, G.~Ding, and J.~Sun, ``Repvgg: Making
  vgg-style convnets great again,'' in \emph{CVPR}, 2021, pp. 13\,733--13\,742.

\bibitem{RepLKNet}
X.~Ding, X.~Zhang, J.~Han, and G.~Ding, ``Scaling up your kernels to 31x31:
  Revisiting large kernel design in cnns,'' in \emph{CVPR}, 2022, pp.
  11\,963--11\,975.

\bibitem{ImageNet}
J.~Deng, W.~Dong, R.~Socher, L.-J. Li, K.~Li, and L.~Fei-Fei, ``Imagenet: A
  large-scale hierarchical image database,'' in \emph{CVPR}, 2009, pp.
  248--255.

\bibitem{WeizhePami}
W.~Liu, M.~Salzmann, and P.~Fua, ``Counting people by estimating people
  flows,'' \emph{IEEE T-PAMI}, 2021.

\bibitem{AutoScale}
C.~Xu, D.~Liang, Y.~Xu, S.~Bai, W.~Zhan, X.~Bai, and M.~Tomizuka, ``Autoscale:
  Learning to scale for crowd counting,'' \emph{IJCV}, pp. 405--434, 2022.

\bibitem{TinyFaces}
P.~Hu and D.~Ramanan, ``Finding tiny faces,'' in \emph{CVPR}, 2017, pp.
  951--959.

\bibitem{TopoCount}
S.~Abousamra, M.~Hoai, D.~Samaras, and C.~Chen, ``Localization in the crowd
  with topological constraints,'' in \emph{AAAI}, 2021, pp. 872--881.

\bibitem{IIM}
J.~Gao, T.~Han, Y.~Yuan, and Q.~Wang, ``Learning independent instance maps for
  crowd localization,'' \emph{arXiv preprint arXiv:2012.04164}, 2020.

\bibitem{ScopedTeacher}
J.~Wang, J.~Gao, Y.~Yuan, and Q.~Wang, ``Crowd localization from gaussian
  mixture scoped knowledge and scoped teacher,'' \emph{arXiv preprint
  arXiv:2206.05717}, 2022.

\bibitem{Bayes}
Z.~Ma, X.~Wei, X.~Hong, and Y.~Gong, ``Bayesian loss for crowd count estimation
  with point supervision,'' in \emph{ICCV}, 2019, pp. 6142--6151.

\bibitem{DMCount}
B.~Wang, H.~Liu, D.~Samaras, and M.~H. Nguyen, ``Distribution matching for
  crowd counting,'' \emph{NeurIPS}, pp. 1595--1607, 2020.

\bibitem{MAN}
H.~Lin, Z.~Ma, R.~Ji, Y.~Wang, and X.~Hong, ``Boosting crowd counting via
  multifaceted attention,'' in \emph{CVPR}, 2022, pp. 19\,628--19\,637.

\bibitem{TransCrowd}
D.~Liang, X.~Chen, W.~Xu, Y.~Zhou, and X.~Bai, ``Transcrowd: Weakly-supervised
  crowd counting with transformers,'' \emph{Science China Information
  Sciences}, pp. 1--14, 2022.

\bibitem{CrowdFormer}
S.~S. Savner and V.~Kanhangad, ``Crowdformer: Weakly-supervised crowd counting
  with improved generalizability,'' \emph{arXiv preprint arXiv:2203.03768},
  2022.

\bibitem{CrowdMLP}
M.~Wang, J.~Zhou, H.~Cai, and M.~Gong, ``Crowdmlp: Weakly-supervised crowd
  counting via multi-granularity mlp,'' \emph{arXiv preprint arXiv:2203.08219},
  2022.

\bibitem{FamNet}
V.~Ranjan, U.~Sharma, T.~Nguyen, and M.~Hoai, ``Learning to count everything,''
  in \emph{CVPR}, 2021, pp. 3394--3403.

\bibitem{MCNN}
Y.~Zhang, D.~Zhou, S.~Chen, S.~Gao, and Y.~Ma, ``Single-image crowd counting
  via multi-column convolutional neural network,'' in \emph{CVPR}, 2016, pp.
  589--597.

\bibitem{QNRF}
H.~Idrees, M.~Tayyab, K.~Athrey, D.~Zhang, S.~Al-Maadeed, N.~Rajpoot, and
  M.~Shah, ``Composition loss for counting, density map estimation and
  localization in dense crowds,'' in \emph{ECCV}, 2018, pp. 532--546.

\bibitem{JHU}
V.~A. Sindagi, R.~Yasarla, and V.~M. Patel, ``Pushing the frontiers of
  unconstrained crowd counting: New dataset and benchmark method,'' in
  \emph{ICCV}, 2019, pp. 1221--1231.

\bibitem{NWPU}
Q.~Wang, J.~Gao, W.~Lin, and X.~Li, ``Nwpu-crowd: A large-scale benchmark for
  crowd counting and localization,'' \emph{IEEE T-PAMI}, pp. 2141--2149, 2020.

\bibitem{c3f}
J.~Gao, W.~Lin, B.~Zhao, D.~Wang, C.~Gao, and J.~Wen, ``C3 framework: An
  open-source pytorch code for crowd counting,'' \emph{arXiv preprint
  arXiv:1907.02724}, 2019.

\bibitem{fpn}
T.-Y. Lin, P.~Doll{\'a}r, R.~Girshick, K.~He, B.~Hariharan, and S.~Belongie,
  ``Feature pyramid networks for object detection,'' in \emph{CVPR}, 2017, pp.
  2117--2125.

\bibitem{adam}
D.~P. Kingma and J.~Ba, ``Adam: A method for stochastic optimization,''
  \emph{arXiv preprint arXiv:1412.6980}, 2014.

\bibitem{Sorting}
Y.~Yang, G.~Li, Z.~Wu, L.~Su, Q.~Huang, and N.~Sebe, ``Weakly-supervised crowd
  counting learns from sorting rather than locations,'' in \emph{ECCV}, 2020,
  pp. 1--17.

\bibitem{MATT}
Y.~Lei, Y.~Liu, P.~Zhang, and L.~Liu, ``Towards using count-level weak
  supervision for crowd counting,'' \emph{Pattern Recognition}, p. 107616,
  2021.

\bibitem{GLCLoss}
X.~Chen and H.~Lu, ``Reinforcing local feature representation for
  weakly-supervised dense crowd counting,'' \emph{arXiv preprint
  arXiv:2202.10681}, 2022.

\bibitem{VGG}
K.~Simonyan and A.~Zisserman, ``Very deep convolutional networks for
  large-scale image recognition,'' \emph{arXiv preprint arXiv:1409.1556}, 2014.

\bibitem{FR}
B.~Kang, Z.~Liu, X.~Wang, F.~Yu, J.~Feng, and T.~Darrell, ``Few-shot object
  detection via feature reweighting,'' in \emph{ICCV}, 2019, pp. 8420--8429.

\bibitem{FSOD}
Q.~Fan, W.~Zhuo, C.-K. Tang, and Y.-W. Tai, ``Few-shot object detection with
  attention-rpn and multi-relation detector,'' in \emph{CVPR}, 2020, pp.
  4013--4022.

\bibitem{GMN}
E.~Lu, W.~Xie, and A.~Zisserman, ``Class-agnostic counting,'' in \emph{ACCV},
  2018, pp. 669--684.

\bibitem{MAML}
C.~Finn, P.~Abbeel, and S.~Levine, ``Model-agnostic meta-learning for fast
  adaptation of deep networks,'' in \emph{ICML}, 2017, pp. 1126--1135.

\end{thebibliography}
}

\end{document}